\pdfoutput=1

\documentclass[11pt,a4paper]{article}
\usepackage[hyperref]{acl2023}
\usepackage{times}
\usepackage{inconsolata}
\usepackage{latexsym}
\usepackage{danudefs}
\usepackage{algorithmic}
\usepackage{algorithm}
\usepackage{color}
\usepackage{booktabs}
\usepackage{xspace}
\usepackage{url}
\usepackage{braket}
\usepackage{multirow}
\usepackage{graphicx}
\usepackage{microtype}

\usepackage[english]{babel}
\usepackage{hyperref}
\addto\captionsenglish{%
}
\addto\extrasenglish{%
}

\usepackage{acronym}
\acrodef{MLM}[MLM]{Masked Language Model}
\acrodef{SoTA}[SoTA]{state-of-the-art}

\usepackage{todonotes}

\makeatletter
\if@todonotes@disabled

\else

\fi
\makeatother

\title{Learning to Predict Concept Ordering for Common Sense Generation}

\author{Tianhui Zhang \And Danushka Bollegala \\
  University of Liverpool, Amazon \\ 
  {\tt \{tianhui.zhang, danushka, bei.peng\}@liverpool.ac.uk} \\ \And Bei Peng}

\date{}

\begin{document}
\maketitle
\begin{abstract}
Prior work has shown that the ordering in which concepts are shown to a commonsense generator plays an important role, affecting the quality of the generated sentence. 
However, it remains a challenge to determine the optimal ordering of a given set of concepts such that a natural sentence covering all the concepts could be generated from a pretrained generator.
To understand the relationship between the ordering of the input concepts and the quality of the generated sentences, we conduct a systematic study considering multiple language models (LMs) and concept ordering strategies.
We find that \texttt{BART-large} model consistently outperforms all other LMs considered in this study when fine-tuned using the ordering of concepts as they appear in CommonGen training data  as measured using multiple evaluation metrics.
Moreover, the larger GPT3-based large language models (LLMs) variants do not necessarily outperform much smaller LMs on this task, even when fine-tuned on task-specific training data.
Interestingly, human annotators significantly reorder input concept sets when manually writing sentences covering those concepts, and this ordering provides the best sentence generations independently of the LM used for the generation, outperforming a probabilistic concept ordering baseline.\footnote{Code: \url{https://github.com/TianhuiZhang/concept_ordering}}
\end{abstract}

\section{Introduction}
\label{sec:intro}


In Generative Commonsense Reasoning (GCR)~\cite{CommonGen}, the goal is to generate a natural commonsense adhering sentence that covers all of the input concepts.
GCR is a challenging natural language generation (NLG) task because it requires (a) relational reasoning using background commonsense knowledge associated with the input concepts, and (b) compositional generalisation ability to work on unseen concept combinations.
CommonGen dataset~\cite{CommonGen} was specifically developed to evaluate the ability of a generative model to produce sentences covering a given set of concepts.
For example, given the set of concepts \{\emph{ball}, \emph{batter}, \emph{pitcher}, \emph{throw}\}, a human annotator would come up with a sentence such as \emph{The \textbf{pitcher} \textbf{throw}s the \textbf{ball}, and the \textbf{batter} hits a home run!}
The input concepts (shown in boldface fonts) are ordered in a specific manner in the produced sentence such as to create a commonsense bearing sentence.

Although neural generation systems produce fluent texts when compared to template-based methods, they fall short in fluency and faithfulness to the input and do not allow control over the output structure~\cite{Puzikov:2018}.
Similar observations are made by prior work on GCR, where the input ordering of concepts has been reported to influence the quality of the generated sentences~\cite{Zhao:2022,Yang:2023}.
However, the relationship between (a) the ordering of the concepts given as the input to a neural text generation model and (b) the architecture of the underlying LM used in the generator remains unexplored, which is the focus of our study in this paper.
Although we consider the concept ordering problem in the context of GCR, we note that it has broader relevance to other tasks in NLP such as multi-document summarisation~\cite{bollegala-etal-2005-machine,bollegala-etal-2006-bottom,Bollegala:IS:2012} and textual coherence modelling~\cite{barzilay-lapata-2005-modeling}

We evaluate the quality of the sentences generated by five commonsense generation LMs: \href{https://github.com/microsoft/unilm}{\texttt{BERT-gen}}~\cite{unilmv2:2020}, \href{https://huggingface.co/facebook/bart-base}{\texttt{BART-base}}~\cite{BART:2019}, \href{https://huggingface.co/t5-base}{\texttt{T5-base}}~\citep{T5:2019}, \href{https://huggingface.co/facebook/bart-large}{\texttt{BART-large}}~\citep{BART:2019} and \href{https://huggingface.co/t5-large}{\texttt{T5-large}}~\citep{T5:2019},
where the input concepts are ordered following three different strategies as described in \autoref{sec:order-methods}.
We use seven evaluation metrics to compare the generated sentences against human-written sentences in CommonGen.
Specifically, we fine-tune each LM on the train sentences in CommonGen dataset using next token prediction as the training objective.

We find \texttt{BART-large} to consistently outperform the other LMs across several evaluation metrics when fine-tuned using the ordering of concepts as they appear in CommonGen training data.
To further compare the concept ordering in the model-generated sentence against the human-generated sentence for the same set of concepts, we use the Kendall rank correlation coefficient (i.e., Kendall's $\tau$) as an evaluation metric.
We find that there exists only a weak correlation ($\tau = 0.328$) between the ordering of concepts shown to the human annotators and the ordering in which those concepts appear in the sentences written by those annotators.
This indicates that human annotators significantly reorder the input concepts when writing sentences that convey commonsense relationships among given concepts.
Interestingly, all the generator models we compare in this paper are able to produce better quality sentences when they are presented with the input concepts in the same ordering as ordered in the sentences.

\section{Related Work}
\label{sec:relate}

\paragraph{Concept Ordering.} Concept Ordering aims to reorder the given concepts in a sequence according to their importance and inner relevance.  Recent work by \citet{Ou:2022} highlighted the effectiveness of pre-trained language models, such as BART~\cite{BART:2019}, in tasks related to concept ordering. \citet{huang:2023} fine-tune the LM to find the most complementary concepts to the given one. \citet{hoyle:2021} and \citep{Zhao:2022} show that suitable concept ordering could increase the quality of the generated sentences. In this work, we aim to explore the relation between the concept ordering and different LMs, assessing their performance in the GCR task.

\paragraph{Generative Commonsense Reasoning.} Recently, a series of works have been proposed to evaluate the commonsense reasoning quality of the model's generation. One strand of research leverages these generations as the external commonsense explanation ~\cite{chen:2023} or chain-of-thought~\cite{zhang:2023}, aiding models in various tasks, including question answering.  Another approach compares the generation with the references, examining if the model could write as natural as human, such as CommonGen~\cite{CommonGen} and ROCStories~\cite{rocstories}. \citet{kgbart:2021} and \citet{Kgr4:2022} incorporate external knowledge into pre-trained LMs to enrich the generation information. 

\paragraph{Text Generation.} Prior work on text generation from structured data such as RDF has shown that the ordering in which a set of entities shown to neural text generation models significantly influences the quality of the generated text~\cite{Moryossef:2019,Zhao:2020}.
Unlike template-based generation methods~\cite{Reiter:2000,Gatt:2018}, neural text generation methods such as Seq2Seq~\cite{Sutskever:NIPS:2014} models perform \emph{text planning} (how to order the inputs) as well as \emph{plan realisation} (how to verbalise the plan) as a single end-to-end task~\cite{Gardent:2017}.

\section{Methods}
\label{sec:task}


\subsection{Task definition}
\label{sec:task-def}
Given a set $\cX = \{x_1, x_2, \ldots, x_m\}$ of lemmatised tokens representing concepts $x_i$, the goal of the GCR task is to generate a natural and grammatically correct sentence $\vec{y} = y_1, y_2, \ldots, y_n$, with tokens (or sub-tokens) $y_j$.
Note that $\cX$ is an \emph{unordered} set of concepts, while the tokens in $\vec{y}$ are ordered.
We must use \emph{all} concepts in $\cX$ when generating $\vec{y}$.
However, we are allowed to use different morphological forms (inflections) of the concepts for this purpose.
Typically in CommonGen dataset ca. $m = 5$, whereas $n$ is determined by the generator model used to produce $\vec{y}$.
Given a set of concepts $\cX$, there exists $m!$ number of possible ordering of concepts.

Let us denote one such ordering by $\vec{x} = x_1, x_2, \ldots, x_m$.
Given $\vec{x}$, a generator parameterised by $\vec{\theta}$ produces the output sequence $\vec{y} = y_1, y_2, \ldots y_n$ according to the generation probabilities given by \eqref{eq:gen-prob}.
\begin{align}
    \label{eq:gen-prob}
    p(\vec{y} | \vec{x}; \vec{\theta}) = \prod_{i=2}^{n} p(y_i | y_{1:i-1}, \vec{x}; \vec{\theta})
\end{align}
We generate sentences $\vec{y}$ using different pre-trained generators for the same set of concepts $\cX$, ordered using different strategies as described in \autoref{sec:order-methods}.

\subsection{Concept Ordering Strategies}
\label{sec:order-methods}
We name the ordering of input concepts in a train/test instance in CommonGen as the \textbf{Original} ordering.
We propose three different strategies to re-order the Original ordering for the purpose of fine-tuning LMs using the CommonGen training instances.
In \textbf{Random} ordering, the concepts in the input set are randomly ordered.
The \textbf{Example} ordering considers the ordering of concepts as they appear in a human-written example sentence in the CommonGen train dataset.\footnote{As detailed in \autoref{sec:format}, in our preliminary experiments, we evaluated three different input formats for a concept ordering:
(a) space delimited, (b) comma delimited, and (c) space delimited with a special end-of-ordering token (i.e., [ORDERING]). 
We found the input format method (c) to perform slightly better than the other two, albeit no significant improvements were observed.}
However, this concept ordering information is not available at test time, we thus use Random Ordering of input concept sets at test time even for the LMs fine-tuned using the Example Ordering.

We induce a \textbf{Probabilistic} ordering among the concepts in a given set using transition probabilities $p(x_j|x_i)$ for ordering the concept $x_j$ after $x_i$ because we want to find a method that could be used both at training and test times and do not extract the concept ordering from pre-trained models' outputs. Therefore, inspired by Glove word embeddings~\cite{glove}, which use the co-occurrence information between words, we use the co-occurrence frequency of each concept pair inside the paths of the ConceptNet to determine the ordering in a given concept set. 
Specifically, we first perform random walks over the ConceptNet graph starting from vertices that appear in the CommonGen train sentences, limiting to a maximum path length of five concepts.
Next, we count the number of paths, $\#(x_i \rightarrow x_j)$, where $x_i$ appears before $x_j$.
The transition probabilities are estimated from the path counts as in \eqref{eq:prob}.
\begin{align}
    \label{eq:prob}
    p(x_j|x_i) = \frac{\#(x_i \rightarrow x_j)}{\#(x_i \rightarrow x_j) + \#(x_j \rightarrow x_i)}
\end{align}
Finally, the probability of generating an ordering $\vec{x} = x_1, x_2, \ldots, x_m$ is computed assuming a first-order Markov chain such that $p(\vec{x}) = \prod_{i=2}^{m} p(x_i|x_{i-1})$.
The \textbf{Probabilistic} ordering strategy enables us to incorporate external knowledge from ConceptNet for the purpose of determining the ordering of input concepts.
Moreover, unlike the \textbf{Example} ordering, \textbf{Probabilistic} ordering can be used both at training and test times.

\section{Experiments}
\label{sec:exp}
\begin{table*}[ht]
\centering
\resizebox{0.95\textwidth}{!}{%
\begin{tabular}{@{}l|l|rrrrrrrr@{}}
\toprule
 & \textbf{Model} & \textbf{BLEU3} & \textbf{BLEU4} & \textbf{ROUGE-2} & \textbf{ROUGE-L} & \textbf{METEOR} & \textbf{CIDEr} & \textbf{SPICE} & \textbf{Coverage} \\ 
\midrule
\multicolumn{1}{c|}{\multirow{4}{*}{\parbox{3cm}{\centering\textbf{Original Ordering~\cite{CommonGen}} \\}}} 
& BERT-Gen & 30.4 & 21.1 & 18.1 & 40.5 & 27.3 & 12.5 & 27.3 & 86.1 \\
\multicolumn{1}{c|}{} & T5-base~ & 26.0 & 16.4 & 14.6 & 34.6 & 23.0 & 9.2 & 22.0 & 76.7 \\
\multicolumn{1}{c|}{} & BART-large~ & 36.3 & 26.3 & \textit{22.2} & 42.0 & \textit{30.9} & 13.9 & 30.6 & \textit{97.4} \\
\multicolumn{1}{c|}{} & T5-large~ & \textit{39.0} & \textit{28.6} & 22.0 & \textit{43.0} & 30.1 & \textit{15.0} & \textit{31.6} & 95.3 \\ 
 \midrule
\multirow{5}{*}{\parbox{3cm}{\centering\textbf{Random} \\ \textbf{ordering}}} 
 & BERT-Gen & 30.5 & 21.1 & 18.0 & 40.1 & 27.4 & 12.6 & 27.4 & 86.5 \\
 & BART-base & 36.0 & 26.6 & 22.0 & 43.3 & 28.8 & 13.9 & 28.5 & 92.0 \\
 & T5-base & 37.7 & 26.9 & 20.9 & 43.2 & 30.2 & 15.1 & 31.2 & 94.2 \\
 & BART-large & \textit{42.4} & \textit{31.8} & \textit{23.8} & 44.7 & \textit{32.8} & \textit{16.6} & \textit{32.3} & \textit{98.8} \\
 & T5-large & 40.6 & 30.0 & 23.1 & \textit{44.9} & 31.5 & 15.9 & 32.1 & 97.6 \\ 
 \midrule
\multirow{5}{*}{\parbox{3cm}{\centering\textbf{Probabilistic} \\ \textbf{ordering}}} 
 & BERT-Gen & 29.1 & 19.9 & 16.9 & 38.6 & 26.7 & 12.2  & 26.7 & 85.7 \\
 & BART-base & 33.5 & 23.8 & 20.4 & 41.2 & 28.0 & 13.4 & 27.8 & 92.6 \\
 & T5-base & 37.6 & 26.8 & 20.5 & 42.4 & 30.6 & 15.1 & 31.1 & 96.4 \\
 & BART-large & 38.5 & 28.0 & 21.8 & 42.4 & \textit{31.7} & 15.4 & 32.2 & \textit{98.8} \\
 & T5-large & \textit{40.8} & \textit{29.8} & \textit{22.2} & \textit{43.9} & 31.6 & \textit{16.2} & \textit{32.4} & 95.7 \\ 
 \midrule
\multirow{5}{*}{\parbox{3cm}{\centering\textbf{Example}\\ \textbf{ordering}}} 
 & BERT-Gen & 33.0 & 23.6 & 19.3 & 41.8 & 28.9 & 13.7 & 28.8 & 93.3 \\
 & BART-base & 36.0 & 26.8 & 22.7 & 44.4 & 29.3 & 14.7 & 29.1 & 97.5 \\
 & T5-base & 40.0 & 30.0 & 22.7 & 45.0 & 31.3 & 16.0 & 32.3 & 97.6 \\
 & BART-large & \textit{\textbf{44.3}} & \textit{\textbf{33.8}} & \textit{\textbf{24.5}} & \textit{\textbf{45.6}} & \textit{\textbf{32.8}} & \textit{\textbf{17.2}} & 33.1 & \textit{\textbf{99.1}} \\
 & T5-large & 43.4 & 32.7 & 23.8 & 45.6 & 32.3 & 16.9 & \textit{\textbf{33.5}} & 98.0 \\ 
 \midrule \midrule
 \multicolumn{1}{l|}{\multirow{5}{*}{\parbox{3cm}{\centering\textbf{Other} \\ \textbf{Prior Methods}}}} 
 & KG-BART~ & 42.1 & 30.9 & 23.4 & 44.5 & 32.9 & 17.5 & 32.7 & 98.7 \\
 & EKI-BART$D_{out}$ & 42.9 & 26.3 & 24.4 & 45.4 & 32.0 & 16.8 & 32.5 & - \\
 & CALM  & - & 29.5 & - & - & 31.9 & 15.6 & 33.2 & - \\
 & NeuroLogic & 41.3 & 36 & - & 44.7 & 31.0 & 15.9 & 31.1 & - \\
 & [MASK] & 43.3 & 32.5 & 24.2 & 44.9 & 32.5 & 17.1 & 32.8 & - \\ 
 \bottomrule
\end{tabular}
}
\caption{Evaluating the quality of the generated sentences on CommonGen test instances.
The input concept sets are ordered using the three strategies described in \autoref{sec:order-methods} during fine-tuning different LMs. Best performing LM fine-tuned with each ordering strategy is shown in italics, while the overall best is shown in bold.}
\label{tab:mainTable}
\end{table*}

\subsection{Experiments setting}
\label{sec:settings}
\paragraph{Dataset:} Our experiments are conducted on the CommonGen dataset~\cite{CommonGen}, which contains 3.5K distinct concept sets (32651/993/1497) with 67389/4018/6042 human written sentences in the training/development/test splits. 
Each instance contains ca. 3-5  input concepts with multiple human-written reference sentences. 

\paragraph{Evaluation Metrics:} We conduct two types of evaluations. First, to evaluate the quality of the generated sentences, we employ the following generation evaluation metrics: BLEU~\cite{bleu}, , ROUGE~\cite{rouge}, METEOR~\cite{meteor}, CIDEr~\cite{cider} and SPICE~\cite{spice}. We report the Coverage of concepts~\cite{CommonGen}, which is defined as the average percentage of input concepts that are present in the generated sentences. Second, we evaluate the ordering of the concepts produced by different LMs (in the order that they appear in the generated sentences) by comparing that to the ordering in human-written test example sentences using the Kendall rank correlation coefficient ($\tau \in [-1,1]$). 
A higher $\tau$ indicates a closer correlation between a concept ordering and their order observed in reference sentences. 
If there are multiple human-written sentences with different orderings for the same input concept set, we take the highest $\tau$ over all of the orderings.

\subsection{Results}
\label{sec:result}
\paragraph{Main Results:} 
We use each concept ordering method to fine-tune five LMs: BERT-Gen~\cite{unilmv2:2020}, T5-base/large~\cite{T5:2019}, and BART-base/large~\cite{BART:2019}.
Implementation details and hyperparameters can be found in \autoref{sec:hyperparameters}.
To assess the effectiveness of the concept set ordering strategies, we compare their performance against two types of prior methods.
The first uses the \textbf{Original} ordering specified in CommonGen~\citep{CommonGen} to fine-tune LMs. 
The second set includes models that incorporate external knowledge, such as KG-BART~\cite{kgbart:2021} and EKI-BART$D_{out}$~\cite{Ekibart:2020}, as well as models that improve performance through pre-training tasks, such as CALM~\cite{CALM:2021}, NeuroLogic~\cite{Neurologic:2021}, and the [MASK]~\cite{Yang:2023}. 

As shown in \autoref{tab:mainTable}, both \textbf{Probabilistic}  and \textbf{Example} ordering outperform the \textbf{Original} Ordering presented in CommonGen~\cite{CommonGen}. 
This shows that refining the ordering strategy for input concepts could enhance an LM's performance in commonsense generation. 
Moreover, \textbf{Example} ordering reports the best performance for all five LMs, even outperforming prior methods that use pre-training tasks or external resources. 

\begin{table}[ht]
\centering
\resizebox{\columnwidth}{!}{%
\begin{tabular}{@{}l|rrrrrr@{}}
\toprule
Ordering & $\tau$ & \textbf{BLEU4} & \textbf{ROUGE-L} & \textbf{METEOR} & \textbf{CIDEr} & \textbf{SPICE} \\ 
 \midrule
Original & 0.328 & 19.0 & 36.4 & 29.0 & 12.9 & 27.7 \\
Random & 0.327 & 18.9 & 36.2 & 29.0 & 12.7 & 27.7 \\
Probabilistic & 0.402 & 20.9 & 37.9 & 29.6 & 13.4 & 29.1 \\ 
\midrule
BERT-Gen & 0.595 & 29.0 & 43.1 & 31.4 & 16.0 & 31.3 \\
BART-base & 0.648 & 31.5 & 44.2 & 31.9 & 16.6 & 32.1 \\
T5-base & 0.627 & 30.3 & 43.9 & 31.9 & 16.4 & 32.0 \\
BART-large & \textbf{0.697} & \textbf{33.4} & \textbf{45.5} & \textbf{32.7} & \textbf{17.3} & \textbf{33.0} \\
T5-large & 0.696 & 32.9 & 45.2 & 32.6 & 17.0 & 32.7 \\
\midrule
Reference & 1 & 35.3 & 57.4 & 33.3 & 18.1 & 33.5\\
\bottomrule
\end{tabular}
}
\caption{Kendall's $\tau$ between the reference ordering and the concept orderings produced using different methods: Lines 2-4 show Original, Random and Probabilistic orderings (\autoref{sec:order-methods}), Lines 5-9 show the orderings of concepts as they appear in the sentences generated using LMs fine-tuned on CommonGen, and Line 10 is the Reference ordering (i.e., ordering of the concepts in human-written CommonGen test sentences).}
\label{tab:diffConceptOrdering}
\end{table}

\begin{table*}[t]
\centering
\resizebox{\textwidth}{!}{%
\begin{tabular}{@{}ll@{}}
\toprule
concept words & throw daughter stream rock daddy \\ \midrule
Reference & Daddy and daughter throwing rocks into stream \\
Turbo(zero shot) & As we hiked beside the stream, my daughter asked if she could throw a rock in the water, but I reminded her that it was too dangerous and daddy \\
Babbage(Original ordering) & little girl throwing a tantrum in the park with her father and sister by the stream and crying because she did not get a toy in the stream. \\
Curie(Original ordering) & mother throwing her daughter in the stream near the rocks with daddy \\
Babbage(Example ordering) & A father and daughter are throwing rocks into a stream. \\
Curie(Example ordering) & daddy and daughter throwing rocks into a stream \\ \midrule
concept words & dog throw frisbee catch \\ \midrule
Reference & A man throws away his dog's favorite frisbee expecting him to catch it in the air. \\
Turbo(zero shot) & The dog loves to play fetch and is always ready to catch the frisbee when it's thrown. \\
Babbage(Original ordering) & A dog is throwing a frisbee. \\
Curie(Original ordering) & A dog is throwing a frisbee and catching it. \\
Babbage(Example ordering) & A dog catching a frisbee and throwing it. \\
Curie(Example ordering) & A dog catching a frisbee thrown by a man. \\ \midrule
concept words & hang squeeze shut head eye \\ \midrule
Reference & Hanging her head, someone squeezes her eyes shut. \\
Turbo(zero shot) & I tried to hang the picture but couldn't do it with just one hand, so I had to squeeze my eyes shut and use both hands to get it done \\
Babbage(Original ordering) & Someone grabs someone by the eyes and squeezes them shut, then drags him away. \\
Curie(Original ordering) & A woman is hanging upside down and squeezing her legs shut, then she opens her eyes. \\
Babbage(Example ordering) & Someone squeezes his eyes shut and hangs his head. \\
Curie(Example ordering) & Someone squeezes his eyes shut and hangs his head. \\ \midrule
concept words & mirror gear picture hold take \\ \midrule
Reference & The man holds the gear and uses the picture taken by the mirror. \\
Turbo(zero shot) & She carefully held the picture up to the mirror to take a closer look at the intricate gear design. \\
Babbage(Original ordering) & A picture of a man in a space suit, holding a mirror, and taking off his gear. \\
Curie(Original ordering) & The mirror has gears in it and a picture of a train engine held up. \\
Babbage(Example ordering) & A man is holding on to a railing to take a picture of his gear and mirror. \\
Curie(Example ordering) & A man is holding a camera and taking a picture of himself in the mirror. \\ \bottomrule
\end{tabular}
}
\caption{Generated examples by different LLMs}
\label{tab:LLMSample}
\end{table*}

\paragraph{Effect of concept ordering:} 
The ordering of concepts as they appear in the human-written CommonGen test sentences (here onwards referred to as the \textbf{Reference} ordering) can be considered as a gold standard for concept ordering.
To evaluate the level of re-ordering of input concepts conducted by an LM, we compare the ordering of concepts in a sentence generated using an LM against their \textbf{Reference} ordering using the Kendall's $\tau$.
Specifically, we first fine-tune each LM using the concept sets as ordered in the CommonGen train instances (i.e., \textbf{Original} ordering).
Next, we present each fine-tuned LM the concept sets in CommonGen test instances in their \textbf{Original} ordering and generate a sentence containing all of the input concepts.
Finally, we compare the ordering of the concepts in the generated sentence against their \textbf{Reference} ordering using $\tau$. The reference sentences in the CommonGen dataset are written by human annotators given the shuffled concepts. Therefore, we think it could be  considered as a “golden” ordering of input concepts. 
To further evaluate the correlation between the ordering of input concepts and the quality of the sentences generated by an LM, we use \texttt{BART-large} fine-tuned with \textbf{Example} ordering as a sentence generator, where we provide it a set of concepts ordered using different methods.


As seen from \autoref{tab:diffConceptOrdering}, among the three concept ordering strategies described in \autoref{sec:order-methods}, the \textbf{Probabilistic} ordering reports the highest $\tau$ value.
In particular, \textbf{Original} has a low $\tau$ value, comparable to \textbf{Random} ordering, indicating that human annotators had to significantly re-order the concept sets shown to them when writing natural sentences.
Interestingly, all five LMs fine-tuned with the \textbf{Example} ordering strategy outperform those three strategies, suggesting that the sentences carry useful contextual clues for determining the ordering of concepts that are exploited by the LMs. We find that the extracted concept ordering generated by BART-large model outperforms that generated by other pre-trained models. It is consistent with the conclusions of \citet{Ou:2022} that a large amount of dependency structure knowledge exists in BART. 
Moreover, we see that $\tau$ values closely aligns with other automatic evaluation metrics, indicating that concept ordering with a higher $\tau$ results in high quality sentence generations. 
Overall, these results suggest that if the concept orderings more closely align with the orderings found in the dataset reference sentences, the Example Ordering trained model could generate superior sentences, leveraging the inherent commonsense knowledge embedded within the pre-trained models.

\subsection{Large Language Model Scenario}
\label{sec:LLM}
Large Language Models (LLMs) have shown impressive performance in complex reasoning compared to the smaller size models~\cite{GPT3:2020, lamda:2022}. 
To investigate their effectiveness for concept ordering, we use the GPT-3 and GPT-3.5~\cite{GPT3:2020} as LLMs and use a prompt-based instruction (see \autoref{sec:LLMdetails} for the details of the prompts) to induce an ordering among a given set of concepts. 
Due to the space limitations we show the results in the Appendix \autoref{tab:LLMEvaluation}.
We find that the Example Ordering outperforms the unordered inputs, indicating that concept ordering is also important to the LLMs. 
Moreover, from the generated sentences (shown in \autoref{tab:LLMSample}) we find that the Example ordering improves their quality as well as the concept coverage. 
Interestingly however, the best results obtained using LLMs with prompts are worse than that compared to the fine-tuned \texttt{BART-large}, which indicates importance of fine-tuning on CommonGen. 

\section{Conclusion}
\label{sec:conclusion}
We examined the impact of concept ordering on the quality of generated sentences in GCR using multiple LMs.
We find that ordering the input concepts can improve performance, and all fine-tuned LMs generate better quality sentences when the input concepts were presented in an order consistent with that found in human-written sentences. 

\section{Limitations}
\label{sec:limit}
In this work, we limited our investigation to the generation of English sentences and to a finite set of pre-trained language models (PLMs).  
This limitation was largely due to the nature of the CommonGen dataset, the only publicly available dataset we found for concept-to-sentence tasks, which is primarily English-centric. 
Therefore, our evaluation of the generation quality was limited to English, which is a morphologically limited language.
Different languages have different grammars and sentence structures, but the automatic evaluation metrics such as BLEU~\cite{bleu}, ROUGE~\cite{rouge} and CIDEr~\cite{cider} could also be used for other languages. 
Therefore, we consider it to be an important next step to evaluate the concept ordering strategies described in this work on languages other than English.

Furthermore, we focus only on the experiments around concept ordering in the LLM scenario. 
However, as reported in much prior work, the commonsense knowledge inside the LLMs could be efficiently exploited by intermediate reasoning steps and designed prompts~\cite{wei:2022,zhang:2023}.  We utilised identical prompts for each backbone LLM and did not investigate the potential influence of prompt variations. we observed that some sentences generated by LLMs contradict commonsense. Therefore, generative commonsense reasoning tasks remain a challenge for LLMs. Given the scope of the current paper, research on prompt design and encoding concept ordering into intermediate steps will be pursued in future work.

Given a set of concepts, it is possible to write multiple natural sentences that arrange the input concepts in different orders.
However, in the CommonGen dataset only a small number of human-written sentences are available for a given concept set, which does not cover all possible orderings.
Therefore, the Kendall's $\tau$ values reported in this paper must be taken as a lower bound on the agreement with a human determining the ordering of concepts in a sentence.
Moreover, we it is important to consider other types of rank evaluation metrics in addition to Kendall's $\tau$ for future evaluations.

\section{Ethical Considerations}
\label{ethical}

While we conducted our research primarily on the CommonGen dataset, which to the best of our knowledge does not present any explicit ethical issues, it is essential to acknowledge the potential for social biases in the LMs~\cite{blodgett2021stereotyping}.
One of the pre-trained language models we used in our experiments, BART are significantly prone to prediction errors related to gender bias~\cite{sharma:2021} and we are not evaluating for the biases in the generated sentences here which should be done before LMs are deployed in the downstream NLP applications. 
Given that we are fine-tuning LMs on the CommonGen dataset, some social biases could get amplified during this fine-tuning process.
The predicted sentences are possibly influenced by such biases. 
It is still an open question for how to effectively mitigate these biases, particularly in the context of generative commonsense reasoning tasks.

\bibliography{GCR-order.bib}
\bibliographystyle{acl_natbib}

\appendix
\section*{Supplementary Appendix}
\begin{table*}[ht]
\centering
\begin{tabular}{l|rrrrrrr}
\toprule
 & \textbf{Parameters} & \textbf{BLEU4} &  \textbf{ROUGE-L} & \textbf{METEOR} & \textbf{CIDEr} & \textbf{SPICE}  \\ 
\midrule
Turbo(zero shot) & 135B & 13.9 &  33.5 & 27.2 & 8.7 & 24.0  \\ 
\midrule
Babbage (Original) & 6.7B & 19.7 &  37.7 & 27.1 & 12.0 & 26.7  \\
Curie(Original) & 13B  & 13.5 &  32.6 & 26 & 10.0 & 24.6  \\ 
\midrule
Babbage (Example) & 6.7B  & 23.6 &  41.0 & 29.2 & 13.6 & 29.2  \\
Curie (Example) & 13B  & 25.5 &  41.9 & 30.0 & 14.4 & 30.1  \\
\bottomrule
\end{tabular}
\caption{Evaluation of the effect of different concept ordering strategies on the LLMs. The results show that a good concept ordering strategy could also help the quality of large model generation.}
\label{tab:LLMEvaluation}
\end{table*}
\section{Different Input Formats for Example Ordering}
\label{sec:format}
We conducted an evaluation to examine the impact of input formats on the quality of generation. 
We designed three types of input formats, given a concept set: 
1) the ordered concepts are concatenated together, 
2) the ordered concepts are concatenated together and separated (delimitted) by commas, 
and 3) the unordered concepts are concatenated with the ordered concepts and denoted by a special ``\textit{[ORDERING]}'' tokens. 
An example using the concept set {\emph{ski}, \emph{mountain}, \emph{skier}}, is provided in \autoref{tab:input_examples}. 
As shown in \autoref{tab:diff_inputs}, the performance of the generated sentences does not change significantly with the different input formatting methods.
However, the best performance among the three input formatting methods is achieved when the concepts are concatenated with the ordered concepts around tokens. 
This may be attributed to the fact that, with the tokens and ordered concepts, a model might better comprehend the ordering task and use the syntactic knowledge embedded in the pre-trained models~\cite{Ou:2022}.

\begin{table}[h]
\centering
\begin{tabular}{@{}l@{\hspace{0.5cm}}l@{}}
\toprule
\textbf{Format} & \textbf{Example} \\
\midrule
w/o Comma & skier ski mountain \\
Comma & skier, ski, mountain \\
Token & ski mountain skier \texttt{[ORDERING]} \\
& skier ski mountain \texttt{[ORDERING]} \\
\bottomrule
\end{tabular}
\caption{Three different input formats given a concept set \{\emph{ski}, \emph{mountain}, \emph{skier}\}}
\label{tab:input_examples}
\end{table}

\begin{table}[h]
\centering 
\resizebox{\columnwidth}{!}{%
\begin{tabular}{@{}l|rrrrr@{}}
\toprule
\textbf{Concept set} & \textbf{BLEU4} & \textbf{ROUGE-L} & \textbf{METEOR} & \textbf{CIDEr} & \textbf{SPICE}  \\ 
\midrule
w/o Comma &  33.4 & 45.5 & 32.6 & 17.1 & 32.8  \\
Comma &  32.5 & 45 & 32.4 & 16.8 & 32.6  \\
Token &  33.8 &  45.6 & 32.8 & 17.2 & 33.1  \\ 
\bottomrule
\end{tabular}
}
\caption{The generated evaluation of different inputs formats.}
\label{tab:diff_inputs}
\end{table}

\section{Concept Ordering on LLMs}
\label{sec:LLMdetails}
We also conducted concept ordering experiments using the CommonGen dataset on LLMs containing more than one billion parameters. 
Specifically, we selected GPT3 models (Babbage and Curie) and GPT3.5 Turbo from OpenAI as our backbone models. 
Since the Turbo model does not support fine-tuning, it was evaluated solely in a zero-shot setting. 
We evaluated the quality of the generated outputs using both the Original ordering and Example ordering strategies on the Babbage and Curie models. 

As shown in \autoref{tab:LLMEvaluation}, the Example Ordering strategy also enhanced the performance of LLMs on the CommonGen dataset task. 
However, we observed that the three LLMs underperform compared against the smaller fine-tuned LMs. 
We notice that this problem also exists in other NLP tasks~\cite{GPT3:2020,gutierrez:2022}. 
Therefore, we examined some sentences generated by these LLMs as presented in \autoref{tab:LLMSample}.
Under the zero-shot setting, Turbo was capable of generating more complex sentences than those provided in the references, and these sentences were consistent with commonsense. When we try to use Example Ordering Strategy for the Turbo model with the prompt \textit{Given a concept list: [concepts], please generate a sentence that aligns the ordering of the concepts:}, we find that the model could not always follow the concept ordering given even under the few-shot setting.

For the Babbage and Curie models, the Original Ordering strategy often resulted in generated sentences that did not encompass all concepts from the given input set. 
The Example Ordering strategy ameliorated this issue; however, even when the grammar was correct, some sentences were inconsistent with commonsense (e.g., \textit{A dog catching a Frisbee and throwing it}). 
We surmise that these issues are the primary reasons why the LLMs did not perform well.
As is suggested by \href{https://platform.openai.com/docs/guides/fine-tuning/preparing-your-dataset}{OpenAI website}, the performance of the model is based on the description given and external content would improve the performance. 
However, the input of CommonGen only has a single concept set. Potential improvements for the performance of LLMs on such tasks could involve providing more content around the concept set such as description about the relations among the concepts.

\section{Training Details}
\label{sec:hyperparameters}
We utilised the BERT, BART, and T5 models as implemented in the Transformers library~\cite{transformers:2020}. Detailed hyper-parameters are provided in the accompanying bash scripts (submitted as supplementary materials).
The primary hyper-parameters were initialised in alignment with the standards set out in the CommonGen paper~\cite{CommonGen}. We fine-tuned each model using the ADAM optimiser~\cite{Adam}. 
To prevent over-fitting, we adopted an early stopping strategy based on the development set's loss. 
We train the model with one NVIDIA A6000 GPU and one V100 GPU. 
Each model could be fine-tuned in less than 6 hours.

The training of the LLMs utilised OpenAI's API.\footnote{\url{https://platform.openai.com/docs/guides}}
Training a Babbage model incurred a cost of \$3 for the CommonGen dataset, while training a Curie model incurred a cost of \$14. During model training, each instance would start with the prompt ``\textit{Generate a sentence containing all the concepts in the concept set:}'' for each concept set and we added a separator `` -\textgreater''  at the end of the prompt. We also use an ending token ``/n'' at the end of the target sentence.
\begin{table}[h]
\centering
\resizebox{\columnwidth}{!}{%
\begin{tabular}{@{}l@{\hspace{0.6cm}}l@{\hspace{0.6cm}}l@{\hspace{0.6cm}}l@{}}
\toprule
\textbf{Models} & \textbf{Random} & \textbf{Probabilistic} & \textbf{Example} \\
\midrule
BERT-Gen & 32*2, 5e-5 & 32*2, 5e-5 & 32*2, 5e-5 \\
BART-base & 64*4, 3e-5 & 64*3, 5e-5 & 64*6, 3e-5 \\
T5-base & 64*4, 5e-5 & 64*3, 5e-5 & 64*4, 5e-5 \\
BART-large & 96*4, 3e-5 & 64*4, 2e-5 & 64*4, 3e-5 \\
T5-large & 32*4, 2e-5 & 64*4, 2e-5 & 64*4, 3e-5 \\
GPT3-babbage & - & - & 128, - \\
GPT3-curie & - & - & 128, - \\
\bottomrule
\end{tabular}%
}
\caption{Key parameters in experiments. In each cell, the left is the batch size and the right is the learning rate.}
\label{tab:keyparameters}
\end{table}


\end{document}